\pdfoutput=1

\documentclass[11pt]{article}

\usepackage[]{acl}

\usepackage{times}
\usepackage{latexsym}

\usepackage[T1]{fontenc}

\usepackage[utf8]{inputenc}

\usepackage{microtype}

\usepackage[utf8]{inputenc} %
\usepackage[T1]{fontenc}    %
\usepackage{hyperref}       %
\usepackage{url}            %
\usepackage{booktabs}       %
\usepackage{amsmath,bm}
\usepackage{amsfonts}
\usepackage{algorithm}
\usepackage{algorithmic}
\usepackage{mathtools}
\usepackage{flushend}
\usepackage{amsmath}
\usepackage{amsfonts}%
\usepackage{nicefrac}       %
\usepackage{xcolor}         %
\usepackage{graphicx}
\usepackage{xspace}
\usepackage{todonotes}
\usepackage{multirow}
\usepackage{wrapfig}
\usepackage{caption}

\usepackage{amsmath,amsfonts,bm}

\def\eqref#1{equation~\ref{#1}}

\def\1{\bm{1}}

\DeclareMathAlphabet{\mathsfit}{\encodingdefault}{\sfdefault}{m}{sl}
\SetMathAlphabet{\mathsfit}{bold}{\encodingdefault}{\sfdefault}{bx}{n}

\newcommand{\ie}{i.e.,\xspace}
\newcommand{\eg}{e.g.,\xspace}
\newcommand{\eat}[1]{}
\newcommand{\paratitle}[1]{\noindent\textbf{#1}\ \ }

\newcommand{\baby}{MetaDistil\xspace}
\newcommand{\citey}[1]{({\citeyear{#1}})}

\title{BERT Learns to Teach: Knowledge Distillation with Meta Learning}

\author{Wangchunshu Zhou$^1$\thanks{\ \ Equal contribution.}\ , Canwen Xu$^{2*}$\thanks{\ \ To whom correspondence should be addressed.},\ \ Julian McAuley$^2$ \\
$^1$Stanford University, $^2$University of California, San Diego \\
$^1$\texttt{wcszhou@stanford.edu}, $^2$\texttt{\{cxu,jmcauley\}@ucsd.edu} \\
}

\begin{document}

\maketitle

\begin{abstract}
We present Knowledge Distillation with Meta Learning (MetaDistil), a simple yet effective alternative to traditional knowledge distillation (KD) methods where the teacher model is fixed during training. We show the teacher network can learn to better transfer 
knowledge to the student network (i.e., \textit{learning to teach}) with the feedback from the performance of the distilled student network in a meta learning framework. Moreover, we introduce a pilot update mechanism to improve the alignment between the inner-learner and meta-learner in meta learning algorithms that focus on an improved inner-learner. Experiments on various benchmarks show that MetaDistil can yield significant improvements compared with traditional KD algorithms and is less sensitive to the choice of different student capacity and hyperparameters, facilitating the use of KD on different tasks and models.\footnote{The code is available at \url{https://github.com/JetRunner/MetaDistil}.}
\end{abstract}

\section{Introduction}
With the prevalence of large neural networks with millions or billions of parameters, model compression is gaining prominence for facilitating efficient, eco-friendly deployment for machine learning applications.
Among 
techniques for compression, knowledge distillation~(KD)~\citep{kd} has shown effectiveness in both Computer Vision and Natural Language Processing tasks~\citep{kd,fitnet,AT,SP,peng2019correlation,vid,RKD,pkt,ab,ft,shi2021learning,sanh2019distilbert,jiao2019tinybert,wang2020minilm}. Previous works often train a large model as the ``teacher''; then they fix the teacher and train a ``student'' model to mimic the behavior of the teacher, in order to transfer the knowledge from the teacher to the student. 

However, this paradigm has the following drawbacks: \textbf{(1) The teacher is unaware of the student's capacity.} Recent studies in pedagogy suggest student-centered learning, which considers students' characteristics and learning capability, has shown effectiveness improving students' performance~\citep{cornelius2007learner,wright2011student}. %
However, in conventional knowledge distillation, the student passively accepts knowledge from the teacher,
without 
regard for the
student model's learning capability and performance. Recent works~\citep{park2021learning,shi2021learning} introduce student-aware distillation by jointly training the teacher and the student with task-specific objectives. However, there is still space for improvement since:
\textbf{(2) The teacher is not optimized for distillation.} In previous works, the teacher is often trained to optimize 
its \textit{own}
inference performance. %
However, the teacher is not aware of 
the need
to transfer its knowledge to a student and thus 
usually does so suboptimally.
A real-world analogy is that a PhD student may have enough knowledge to solve problems themselves, but requires additional teaching training to qualify as a professor.  
\begin{figure*}
  \centering
  \includegraphics[width=\textwidth]{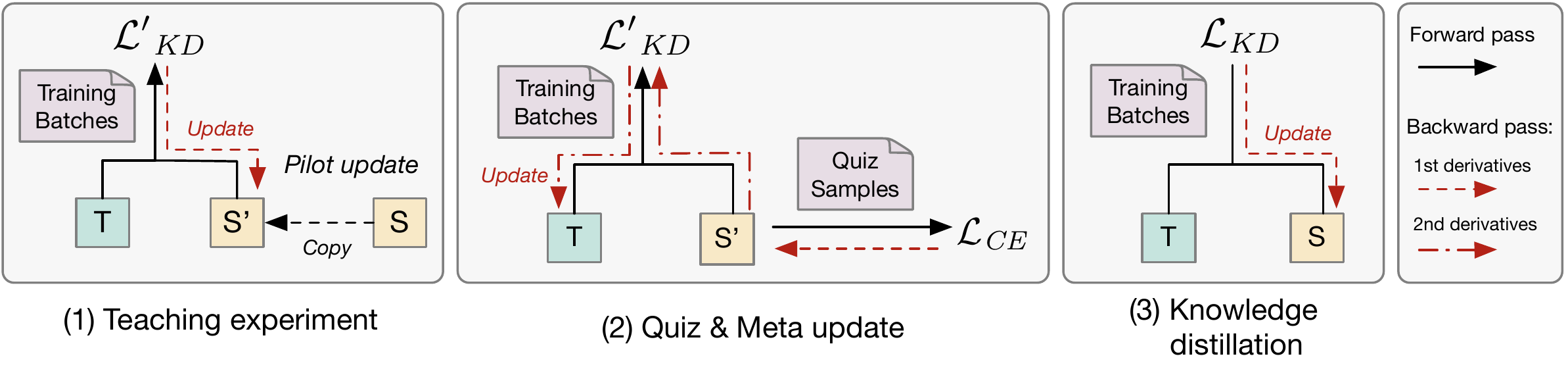}
  \caption{The workflow of \baby. (1) We perform 
  experimental knowledge distillation 
  on %
  a selection of
  training batches. Instead of updating the student $S$, we make a temporary copy $S'$ and update $S'$. (2) We calculate a Cross-Entropy loss $\mathcal{L}_\mathit{CE}$ of $S'$ on samples from a separate quiz set. We calculate the gradients of $\mathcal{L}_\mathit{CE}$ with respect to the parameters of $T$ and update $T$ by gradient descent. (3) We discard $S'$ and use the updated $T$ to perform
  actual knowledge distillation and update $S$. \label{fig:metakd}}
\end{figure*}

To address these two drawbacks, we propose Knowledge Distillation with Meta Learning (\baby), a new teacher-student distillation framework using meta learning~\citep{MAML} to exploit feedback about the student's learning progress to improve the teacher's knowledge transfer ability throughout the distillation process. On the basis of 
previous formulations of bi-level optimization based meta learning~\citep{MAML}, we propose a new mechanism %
called \textit{pilot update} that aligns the learning of the bi-level learners (\ie the teacher and the student). %
We illustrate the workflow of \baby in Figure~\ref{fig:metakd}. The teacher in \baby is trainable, which enables the teacher to adjust to its student network and also improves its ``teaching skills.'' Motivated by the idea of student-centered learning, we allow the teacher to adjust its output based on the performance of the student model on a ``quiz set,'' which is a separate reserved data split from the original training set. For each training step, we first copy the student $S$ to $S'$ and update $S'$ by a common knowledge distillation loss. We call this process 
a
``teaching experiment.'' In this way, we can obtain an experimental student $S'$ that can be 
quizzed.
Then, we sample 
from the quiz set, and calculate 
the loss of
$S'$ on these samples. We use this loss as a feedback signal to meta-update the teacher by calculating second derivatives and performing gradient descent~\citep{MAML}. Finally, we discard the experimental subject $S'$ and use the updated teacher to 
distill into the student $S$ on the same training batches. The use of meta learning allows the teacher model to receive feedback from the student in a completely differentiable way. We provide a simple and intuitive approach to explicitly optimize the teacher using the student's quiz performance as a proxy. 

To test the effectiveness of 
\baby, we conduct extensive experiments on text
and image classification tasks.
\baby outperforms knowledge distillation by a large margin, verifying the effectiveness and versatility of our method. Also, our method achieves state-of-the-art performance compressing BERT~\citep{devlin2018bert} on the GLUE benchmark~\citep{glue} and shows competitive results compressing ResNet~\citep{he2016deep} and VGG~\citep{vgg} on CIFAR-100~\citep{krizhevsky2009learning}. Additionally, we design experiments to analyze and explain the improvement. %
Ablation studies show the
effectiveness of our proposed pilot update and dynamic distillation. Also, compared to conventional KD, \baby is more robust to different student capacity and hyperparameters, which is probably because of its ability to adjust the parameters of the teacher model. %

\section{Related Work}
\paragraph{Knowledge Distillation} Recently, many attempts have been made to accelerate large neural networks~\citep{xu2020bert,xu2021survey,pabee,ssr,xu2022survey}. Knowledge distillation is a 
prominent
method for training compact networks to achieve comparable performance 
to
a deep network. \citet{kd} first introduced the idea of knowledge distillation to exploit the ``dark knowledge'' (i.e., soft label distribution) from a large teacher model as additional supervision for training a smaller student model. Since its introduction, 
several
works~\citep{fitnet,AT,SP,RKD,sun2019patient,jiao2019tinybert} %
have investigated
methods that align different latent representations between the student and
teacher models for better knowledge transfer. In the context of knowledge distillation, \baby shares some common ideas with the line of work that utilizes a sequence of intermediate teacher models to make the teacher network better adapt to the capacity of the student model throughout the training process, including teacher assistant knowledge distillation (TAKD)~\citep{TA} and route constraint optimization (RCO)~\citep{RCO}. However, the intermediate teachers are heuristically selected independently of the training process and the 
evolution
of the teacher network is discrete. 
In contrast, \baby employs meta learning to make the teacher model adapt to the current state of the student model
and provide a continuously evolving meta-teacher that can better teach the student. 
Concurrently, \citet{park2021learning} and \citet{shi2021learning} propose to update the teacher model jointly with the student model with task specific objectives (e.g., cross-entropy loss) during the KD process and add constraints to keep student and teacher similar to each other. Their approaches makes the teacher model aware of the student model by constraining the teacher model's capacity. However, the teacher models in their methods are still not optimized for knowledge transfer. In addition, \citet{DML} introduced deep mutual learning where multiple models learn collaboratively and teach each other throughout the training process. While it is focused on a different setting where different models have approximately the same capacity and are learned from scratch, it also encourages the teacher model to behave similarly to the student model. Different from all aforementioned methods, \baby employs meta learning to explicitly optimize the teacher model for better knowledge transfer ability,
and leads to improved performance of the resulting student model.%

\paragraph{Meta Learning} The core idea of meta learning is ``learning to learn,'' which means taking the optimization process of a learning algorithm into 
consideration 
when
optimizing the learning algorithm itself. Meta learning typically involves a bi-level optimization process where the inner-learner provides feedback for optimization of the meta-learner. Successful applications of meta learning include learning better initialization~\citep{MAML}, architecture search~\citep{DARTS}, learning to optimize the learning rate schedule~\citep{baydin2018online}, and learning to optimize~\citep{andrychowicz2016learning}. These works typically aim to obtain an optimized meta-learner (\ie the teacher model in \baby), while the optimization of the inner-learner (\ie the student model in \baby), is mainly used to provide learning signal for the meta optimization process. This is different from the objective of knowledge distillation where an optimized student model is the goal.
Recently, there have been a few works investigating using this bi-level optimization framework to obtain a better inner-learner. For example, meta pseudo labels~\citep{pham2020meta} use meta learning to optimize a pseudo label generator for better semi-supervised learning; meta back-translation~\citep{pham2021meta} meta-trains a back-translation model to better train a machine translation model. These methods adapt the same bi-level optimization process as previous works where the goal is to obtain an optimized meta-learner. In these approaches, during each iteration, the meta-learner is optimized for the original inner-learner and then applied to the updated inner-learner in the next iteration. This leads to a mismatch between the meta-learner and the inner-learner, and is therefore suboptimal for learning a good inner-learner. In this paper, we introduce a pilot update mechanism, which is a simple and general method for this kind of problems, for the inner-learner to mitigate this issue and make the updated meta-learner better adapted to the inner-learner. 

\paragraph{Meta Knowledge Distillation} %
Recently, some works on KD take a meta approach.
\citet{pan2020meta} proposed a framework to train a meta-teacher across domains that can better fit new domains with meta-learning. Then, traditional KD is performed to transfer the knowledge from the meta-teacher to the student. \citet{liu2020metadistiller} proposed a self-distillation network which utilizes meta-learning to train a label-generator as a fusion of deep layers in the network, to generate more compatible soft targets for shallow layers. Different from the above, \baby is a general knowledge distillation method that exploits meta-learning to allow the teacher to learn to teach dynamically. Instead of merely training a meta-teacher, our method uses meta-learning throughout the procedure of knowledge transfer, making the teacher model compatible for the student model for every training example during each training stage.

\section{Knowledge Distillation with Meta Learning}

An overview of \baby is presented in Figure \ref{fig:metakd}. 
\baby includes two major components. First, the meta update enables the teacher model to receive the student model's feedback on the distillation process, allowing the teacher model to ``learn to teach'' and provide distillation signals that are more suitable for the student model's current capacity. The pilot update mechanism 
ensures a finer-grained match between the student model and the meta-updated teacher model. 

\subsection{Background}

\subsubsection{Knowledge Distillation} Knowledge distillation algorithms aim to exploit the hidden knowledge from a large teacher network, denoted as $T$, to guide the training of a shallow student network, denoted as $S$. To help transfer the knowledge from the teacher to the student, apart from the original task-specific objective (e.g., cross-entropy loss), a knowledge distillation objective which aligns the behavior of the student and the teacher is included to train the student network. Formally, given a labeled dataset $\mathcal{D}$ of $N$ samples $\mathcal{D} = \{\left(x_1, y_1\right), \dots, \left(x_N, y_N\right)\}$, we can write the loss function of the student network as follows,
\begin{equation}
\begin{split}
    \mathcal{L}_{S}\left(\mathcal{D};  \theta_{S}; \theta_{T} \right)= \frac{1}{N} \sum_{i=1}^N [ \alpha\mathcal{L}_{\mathcal{T}}\left(y_i,  S\left(x_i; \theta_{S} \right) \right) \\ + \left(1 - \alpha\right) \mathcal{L}_{\mathit{KD}}\left(T\left(x_i; \theta_{T}\right),S\left(x_i; \theta_{S} \right)\right) ] \label{eq:loss}
\end{split}
\end{equation}
where $\alpha$ is a hyper-parameter to control the relative importance of the two terms; $\theta_{T}$ and $\theta_{S}$ are the parameters of the teacher $T$ and student $S$, respectively. $\mathcal{L}_{\mathcal{T}}$ refers to the task-specific loss 
and $\mathcal{L}_{\mathit{KD}}$ refers to the knowledge distillation loss which measures the similarity of the student and the teacher.
Some popular similarity measurements include the KL divergence between the output probability distribution, the mean squared error (MSE) between student and teacher logits, the similarity between the student and the teacher's attention distribution, etc. We do not specify the detailed form of the loss function because \baby is a general framework that can be easily applied to various kinds of KD objectives as long as the objective is differentiable with respect to the teacher parameters. In the experiments of this paper, we use mean squared error between the hidden states of the teacher and the student for both our method and the KD baseline since recent study~\citet{kim2021comparing} finds that it is more stable and slightly outperforms than KL divergence.

\subsubsection{Meta Learning} In meta learning algorithms that involve a bi-level optimization problem~\citep{MAML}, there exists an inner-learner $f_i$ and a meta-learner $f_m$. The inner-learner is trained to accomplish a task $\mathcal{T}$ or a distribution of tasks with help from the meta-learner. The training process of $f_i$ on $\mathcal{T}$ with the help of $f_m$ is typically called \textit{inner-loop}, and we can denote $f'_i(f_m)$ as the updated inner-learner after the inner-loop. We can express $f'_i$ as a function of $f_m$ because 
learning 
$f_i$ depends on $f_m$. In return, the meta-learner is optimized with a meta objective, which is generally the maximization of expected performance of the inner-learner after the inner-loop, \ie $f'_i(f_m)$. This learning process is called a \emph{meta-loop} and is often accomplished by gradient descent with derivatives of $\mathcal{L}(f'_i(f_m))$, the loss of updated inner-leaner on some held-out support set (\ie the quiz set in our paper).

\subsection{Methodology}
\label{sec:method}
\subsubsection{Pilot Update} \label{sec:pu}
In the original 
formulation of
meta learning~\citep{MAML}, the purpose is to learn a good meta-learner $f_m$ that can generalize to 
different inner-learners $f_i$ for different tasks. In their approach, the meta-learner is optimized for the ``original'' inner-learner at the beginning of each iteration and the current batch of training data. The updated meta-learner is then applied to the updated inner-learner and a different batch of data in the next iteration. This behavior is reasonable if the purpose is to optimize the meta-learner. However, in \baby, we only care about the performance of the only inner-learner, \ie the student. In this case, this behavior leads to a mismatch between the meta-learner and the inner-learner, and is therefore suboptimal for learning a good inner-learner. Therefore, we need a way to align and synchronize the learning of the meta- and inner-learner, in order to allow an update step of the meta-learner to have an instant effect on the inner-learner. This instant reflection prevents the meta-learner from catastrophic forgetting~\citep{mccloskey1989catastrophic}. To achieve this, we design a pilot update mechanism. For a batch of training data $\bm x$, we first make a temporary copy of the inner-learner $f_i$ and update both the copy $f'_i$ and the meta learner $f_m$ on $\bm x$. Then, we discard $f'_i$ and update $f_i$ again with the updated $f_m$ on the same data $\bm x$. This mechanism can apply the impact of data $\bm x$ to both $f_m$ and $f_i$ at the same time, thus aligns the training process. Pilot update is a general technique that can potentially be applied to any meta learning application that optimizes the inner-learner performance. We will describe how we apply this mechanism to \baby 
shortly 
and empirically verify the effectiveness of pilot update in Section~\ref{sec:res}.

\begin{algorithm*}[t]
\small
\caption{Knowledge Distillation with Meta Learning (\baby)}
\label{alg:maml}
\begin{algorithmic}[1]
\REQUIRE student $\theta_S$, teacher $\theta_T$, train set $\mathcal{D}$, quiz set $\mathcal{Q}$
\REQUIRE $\lambda$, $\mu$: learning rate for the student and the teacher %
\WHILE{not done}
\STATE Sample batch of training data $\bm x \sim \mathcal{D}$
 \STATE Copy student parameter $\theta_S$ to student $\theta_S'$
 \STATE Update $\theta_S'$ with $\bm x$ and $\theta_T$: $\theta_S' \leftarrow \theta_S'-\lambda \nabla_{\theta_S'}  \mathcal{L}_S(x; \theta_S; \theta_T)$
 \STATE Sample a batch of quiz data $\bm q \sim \mathcal{Q}$
 \STATE Update $\theta_T$ with $\bm q$ and $\theta_S'$: $\theta_T \leftarrow \theta_T - \mu
 \nabla_{\theta_T} \mathcal{L}_{\mathcal{T}}\left(\bm q, \theta_S'(\theta_T)\right)$
 \STATE Update original $\theta_S$ with $\bm x$ and the updated $\theta_T$: $\theta_S \leftarrow \theta_S-\lambda \nabla_{\theta_S}  \mathcal{L}_S(x; \theta_S; \theta_T)$
\ENDWHILE
\end{algorithmic}
\end{algorithm*}

\begin{table*}[t]
\begin{center}
\resizebox{1\linewidth}{!}{
\begin{tabular}{lcccccccccc}
\toprule
\multirow{3}{*}{\textbf{Method}} & \multirow{3}{*}{\textbf{\#Param.}} & \multirow{3}{*}{\textbf{Speed-up}}  & \multicolumn{8}{c}{\textbf{GLUE}~\citep{glue}} \\
\cmidrule{4-11}
& & & \textbf{CoLA} & \textbf{MNLI} & \textbf{MRPC} & \textbf{QNLI} & \textbf{QQP} & \textbf{RTE} & \textbf{SST-2} & \textbf{STS-B} \\
& & & (8.5K) & (393K) & (3.7K) & (105K) & (364K) & (2.5K) & (67K) & (5.7K) \\
\midrule
\multicolumn{11}{c}{\textbf{Dev Set}} \\
\midrule
BERT-Base (teacher)~\citep{devlin2018bert} & 110M & 1.00$\times$ & 58.9 & 84.6/84.9 & 91.6/87.6 & 91.2 & 88.5/91.4 & 71.4 & 93.0 & 90.2/89.8 \\
BERT-6L (student)~\citep{turc2019well} & 66M & 1.94$\times$ & 53.5 &  81.1/81.7 & 89.2/84.4 & 88.6 & 86.9/90.4 & 67.9 & 91.1 & 88.1/87.9 \\
\midrule
\multicolumn{11}{c}{\textit{Pretraining Distillation}} \\ \midrule
TinyBERT$^\ddagger$~\citep{jiao2019tinybert} & 66M & 1.94$\times$ & 54.0 & 84.5/84.5 & 90.6/86.3 & 91.1 & 88.0/91.1 & 73.4 & 93.0 & 90.1/89.6 \\
MiniLM~\citep{wang2020minilm} & 66M & 1.94$\times$ & 49.2 & 84.0/~~~-~~~  & 88.4/~~~-~~~ & 91.0 & ~~~-~~~/91.0 & 71.5 & 92.0 & - \\
MiniLM v2~\citep{wang2020minilmv2} & 66M & 1.94$\times$ & 52.5 & 84.2/~~~-~~~  & 88.9/~~~-~~~ & 90.8 & ~~~-~~~/91.1 & 72.1 & 92.4 & - \\
\midrule
\multicolumn{11}{c}{\textit{Task-specific Distillation}} \\ \midrule
KD$^\dagger$~\citep{kd} & 66M & 1.94$\times$ & 54.1 & 82.6/83.2 & 89.6/85.2 & 89.2 & 87.3/90.9 & 67.7 & 91.2 & 88.6/88.2  \\
PKD$^\dagger$~\citep{sun2019patient} & 66M & 1.94$\times$ & 54.5 & 82.7/83.3 & 89.4/84.7 & 89.5 & 87.8/90.9 & 67.6 & 91.3 & 88.6/88.1 \\
TinyBERT w/o DA$^\dagger$ & 66M & 1.94$\times$ & 52.4 & \textbf{83.6}/\textbf{83.8} & 90.5/86.5 & 89.8 & 87.6/90.6 & 67.7 & 91.9 & 89.2/88.7 \\
RCO$^\dagger$~\citep{RCO} & 66M & 1.94$\times$ & 53.6 & 82.4/82.9 & 89.5/85.1 & 89.7 & 87.4/90.6 & 67.6 & 91.4 & 88.7/88.3 \\
TAKD$^\dagger$~\citep{TA} & 66M & 1.94$\times$ & 53.8 & 82.5/83.0 & 89.6/85.0 & 89.6 & 87.5/90.7 & 68.5 & 91.4 & 88.2/88.0 \\
DML$^\dagger$~\citep{DML} & 66M & 1.94$\times$ & 53.7 & 82.4/82.9 & 89.6/85.1 & 89.6 & 87.4/90.3 & 68.4 & 91.5 & 88.4/88.1 \\
ProKT$^\dagger$~\citep{shi2021learning} & 66M & 1.94$\times$ & 54.3 & 82.8/83.2 & 90.7/86.3 & 89.7 & 87.9/90.9 & 68.4 & 91.3 & 88.9/88.6 \\
SFTN$^\dagger$~\citep{park2021learning} & 66M & 1.94$\times$ & 53.6 & 82.4/82.9 & 89.8/85.3 & 89.5 & 87.5/90.4 & 68.5 & 91.5 & 88.4/88.5 \\
\baby \textit{(ours)} & 66M & 1.94$\times$ & \bf 58.6 & 83.5/\textbf{83.8} & \bf 91.1/86.8 & \bf 90.4 & \bf 88.1/91.0 & \bf 69.4 & \bf 92.3 & \bf 89.4/89.1 \\
\quad w/o pilot update & 66M & 1.94$\times$ &  56.3 & 83.0/83.4 & 90.6/86.6 & 89.9 & 88.0/88.5 & 67.7 & 92.0 & 89.2/89.0 \\
\midrule
\multicolumn{11}{c}{\textbf{Test Set}} \\
\midrule
BERT-Base (teacher)~\citep{devlin2018bert} & 110M & 1.00$\times$ & 52.1 & 84.6/83.4 & 88.9/84.8 & 90.5 & 71.2/89.2 & 66.4 & 93.5 & 87.1/85.8 \\
\midrule
\multicolumn{11}{c}{\textit{Pretraining Distillation}} \\ \midrule
DistilBERT~\citep{sanh2019distilbert} & 66M & 1.94$\times$ & 45.8 & 81.6/81.3 & 87.6/83.1 & 88.8 & 69.6/88.2 & 54.1 & 92.3 & 71.0/71.0 \\
TinyBERT$^\ddagger$~\citep{jiao2019tinybert} & 66M & 1.94$\times$ & 51.1 & 84.3/83.4 & 88.8/84.5 & 91.6 & 70.5/88.3 & 70.4 & 92.6 & 86.2/84.8 \\
\midrule
\multicolumn{11}{c}{\textit{Task-specific Distillation}} \\ \midrule
KD~\citep{turc2019well} & 66M & 1.94$\times$ & - & 82.8/82.2 & 86.8/81.7 & 88.9 & 70.4/88.9 & 65.3 & 91.8 & - \\
PKD~\citep{sun2019patient} & 66M & 1.94$\times$ & 43.5 & 81.5/81.0 & 85.0/79.9 & 89.0 & 70.7/88.9 & 65.5 & 92.0 & 83.4/81.6 \\ 
BERT-of-Theseus~\citep{xu2020bert} & 66M & 1.94$\times$ & 47.8 & 82.4/82.1 & 87.6/83.2 & 89.6 & \bf 71.6/89.3 & 66.2 & 92.2 & 85.6/84.1 \\
ProKT~\citep{shi2021learning} & 66M & 1.94$\times$ & - & 82.9/82.2 & 87.0/82.3 & 89.7 & 70.9/88.9 & - & 93.3 & - \\
TinyBERT$^\ddagger$~\citep{jiao2019tinybert} & 66M & 1.94$\times$ & 47.5 & 83.0/82.6 & 87.9/82.8 & 89.8 & 70.9/88.6 & 66.8 & 93.1 & 85.8/84.6 \\
DML$^\dagger$~\citep{DML} & 66M & 1.94$\times$ & 48.5 & 82.6/81.6 &  86.5/81.2 & 89.5 & 70.7/88.7 & 66.3 & 92.7 & 85.5/84.0 \\
RCO$^\dagger$~\citep{RCO} & 66M & 1.94$\times$ & 48.2 & 82.3/81.2 &  86.8/81.4 & 89.3 & 70.4/88.7 & 66.5 & 92.6 & 85.3/84.1 \\
TAKD$^\dagger$~\citep{TA} & 66M & 1.94$\times$ & 48.4 & 82.4/81.7 & 86.5/81.3 & 89.4 & 70.6/88.8 & 66.8 & 92.9 & 85.4/84.1 \\
SFTN$^\dagger$~\citep{park2021learning} & 66M & 1.94$\times$ & 48.1 & 82.1/81.3 & 86.5/81.2 & 89.6 & 70.2/88.4 & 66.3 & 92.7 & 85.1/84.2 \\
\baby \textit{(ours)} & 66M & 1.94$\times$ & \bf 50.7 & \bf 83.8/83.2 & \bf 88.7/84.7 & \bf 90.2 & 71.1/88.9 & \bf 67.2 & \bf 93.5 & \bf 86.1/85.0 \\
\quad w/o pilot update & 66M & 1.94$\times$ & 49.1 & 83.3/82.8 & 88.2/84.1 & 89.9 & 71.0/88.7 & 66.6 & 93.5 & 85.9/84.6 \\
\bottomrule
\end{tabular}
}
\end{center}
\caption{Experimental results
on the development set and the test set of GLUE. Numbers under each dataset indicate the number of training samples. All student models have the same architecture of 66M parameters, 6 Transformer layers and 1.94$\times$ speed-up. The test results are from the official test server of GLUE. The best results for the task-specific setting are marked with \textbf{boldface}. Results reported by us are average of 3 runs with different seeds. $^\dagger$Results reported by us. The student is initialized with a 6-layer pretrained BERT~\citep{turc2019well} thus has a \emph{better} performance than the original implementation. $^\ddagger$TinyBERT has data augmentation (DA).} 
\label{tab:main}
\end{table*}

\subsubsection{Learning to Teach} In \baby, we would like to optimize the teacher model, which is fixed in traditional KD frameworks. Different from previous deep mutual learning~\citep{DML} methods that switch the role between the student and teacher network and train the original teacher model with soft labels generated by the student model, or recent works~\citep{shi2021learning,park2021learning} that update the teacher model with a task-specific loss during the KD process, \baby explicitly optimizes the teacher model in a ``learning to teach'' fashion, so that it can better transfer its knowledge to the student model. Concretely, the optimization objective of the teacher model in the \baby framework is \textit{the performance of the student model after distilling from the teacher model.} This ``learning to teach'' paradigm naturally fits the bi-level optimization framework in meta learning literature. 

In the \baby framework, the student network $\theta_S$ is the inner-learner and the teacher network $\theta_T$ is the meta-learner. For each training step, we first copy the student model $\theta_S$ to an ``experimental student'' $\theta_S'$. Then given a batch of training examples $\bm x$ and the learning rate $\lambda$, the experimental student is updated in the same way as conventional KD algorithms:
\begin{equation}
\theta_S'(\theta_T) = \theta_S-\lambda \nabla_{\theta_S}  \mathcal{L}_S(x; \theta_S; \theta_T).
\label{eq:student}
\end{equation}
To simplify notation,
we will consider one
gradient update for the rest of this section, but using multiple gradient updates is a straightforward extension. We observe that the updated experimental student parameter $\theta_S'$, as well as the student quiz loss $l_q = \mathcal{L}_{\mathcal{T}}(\bm q, \theta_S'(\theta_T))$ on a batch of quiz samples $\bm q$ sampled from a held-out quiz set $\mathcal{Q}$, is a function of the teacher parameter $\theta_T$. Therefore, we can optimize $l_{q}$ with respect to $\theta_T$ by a learning rate $\mu$:
\begin{equation}
\label{eqn:meta_update}
\theta_T \leftarrow \theta_T-\mu \nabla_{\theta_T} \mathcal{L}_{\mathit{\mathcal{T}}}\left(\bm q, \theta_S'(\theta_T)\right)%
\end{equation}
We evaluate the performance of the experimental student on a separate quiz set to prevent overfitting the validation set, which is preserved for model selection. Note that the student is never trained on the quiz set and the teacher only performs meta-update on the quiz set instead of fitting it. We do not use a dynamic quiz set strategy because otherwise the student would have been trained on the quiz set and the loss would not be informative. After meta-updating the teacher model, we then update the ``real'' student model in the same way as described in Equation~\ref{eq:student}. 
Intuitively, optimizing the teacher network $\theta_T$ with Equation~\ref{eqn:meta_update} is maximizing the expected performance of the student network after being taught by the teacher with the KD objective in the inner-loop. This meta-objective allows the teacher model to adjust its parameters to better transfer its knowledge to the student model.
We apply the pilot update strategy described in Section~\ref{sec:pu} to better align the learning of the teacher and student, as shown in Algorithm~\ref{alg:maml}.

\section{Experiments}
\label{sec:exp}

\subsection{Experimental Setup}
We evaluate \baby on two commonly used classification benchmarks for knowledge distillation in both Natural Language Processing and Computer Vision (see Appendix~\ref{appendix:cv}).

\paragraph{Settings}
For NLP, we evaluate our proposed approach on the GLUE benchmark~\citep{glue}. Specifically, we test on MRPC~\citep{mrpc}, QQP and STS-B~\citep{senteval} for Paraphrase Similarity Matching; SST-2~\citep{sst} for Sentiment Classification; MNLI~\citep{mnli}, QNLI~\citep{qnli} and RTE~\citep{glue} for the Natural Language Inference; CoLA~\citep{cola} for Linguistic Acceptability.
Following previous studies~\citep{sun2019patient,jiao2019tinybert,xu2020bert}, our goal is to distill BERT-Base~\citep{devlin2018bert} into a 6-layer BERT with the hidden size of 768. We use MSE loss between model logits as the distillation objective. 
The reported results are in the same format as on the GLUE leaderboard.
For MNLI, we report the results on MNLI-m and MNLI-mm, respectively. For MRPC and QQP, we report both F1 and accuracy. For STS-B, we report Pearson and Spearman correlation. The metric for CoLA is Matthew's correlation. The other tasks use accuracy as the metric.

Following previous works~\citep{sun2019patient,turc2019well,xu2020bert}, we evaluate \baby in a \textit{task-specific} setting where the teacher model is fine-tuned on a downstream task and the student model is trained on the task with the KD loss. We do not choose the pretraining distillation setting since it requires significant computational resources. We implement \baby based on Hugging Face Transformers~\citep{hf}. %

\paragraph{Baselines} For comparison, we report the results of vanilla KD and patient knowledge distillation~\citep{sun2019patient}. We also include the results of progressive module replacing~\citep{xu2020bert}, a state-of-the-art task-specific compression method for BERT which also uses a larger teacher model to improve smaller ones like knowledge distillation. In addition, according to~\citet{turc2019well}, the reported performance of current task-specific BERT compression methods is underestimated because the student model is not appropriately initialized. To ensure fair comparison, we re-run task-specific baselines with student models initialized by a pretrained 6-layer BERT model and report our results in addition to the official numbers in the original papers.
We also compare against deep mutual learning (DML)~\citep{DML}, teacher assistant knowledge distillation (TAKD)~\citep{TA}, route constraint optimization (RCO)~\citep{RCO}, proximal knowledge teaching (ProKT)~\citep{shi2021learning}, and student-friendly teacher network (SFTN)~\citep{park2021learning}, where the teacher network is not fixed. For reference, we also present results of pretraining distilled models including DistilBERT~\citep{sanh2019distilbert}, TinyBERT~\citep{jiao2019tinybert}, MiniLM v1 and v2~\citep{wang2020minilm,wang2020minilmv2}. Note that among these baselines, PKD~\citep{sun2019patient} and Theseus~\citep{xu2020bert} exploit intermediate features while TinyBERT and the MiniLM family use both intermediate and Transformer-specific features. In contrast, \baby uses none of these but the vanilla KD loss (Equation~\ref{eq:loss}).

\paragraph{Training Details} For training hyperparameters, we fix the maximum sequence length to 128 and the temperature to 2 for all tasks. For our method and all baselines (except those with officially reported numbers), we perform grid search over the sets of the student learning rate $\lambda$ from \{1e-5, 2e-5, 3e-5\}, the teacher learning rate $\mu$ from \{2e-6, 5e-6, 1e-5\}, the batch size from \{32, 64\}, the weight of KD loss from \{0.4, 0.5, 0.6\}. We randomly split the original training set to a new training set and the quiz set by $9:1$. For RCO, we select four unconverged teacher checkpoints as the intermediate training targets. For TAKD, we use KD to train a teacher assistant model with 10 Transformer layers.

\subsection{Experimental Results}
\label{sec:res}
We report the experimental results on both the development set and test set of the eight GLUE tasks~\citep{glue} in Table~\ref{tab:main}. \baby achieves state-of-the-art performance under the task-specific setting and outperforms all KD baselines. Notably, without using any intermediate or model-specific features in the loss function, \baby outperforms methods with carefully designed features, \eg PKD and TinyBERT (without data augmentation). Compared with other methods with a trainable teacher~\citep{DML,TA,RCO,shi2021learning}, our method still demonstrates 
superior performance.
As we analyze, with the help of meta learning, \baby is able to directly optimize the teacher's teaching ability thus 
yielding
a further improvement in terms of student accuracy. Also, we observe a performance drop by replacing pilot update with a normal update. This ablation study verifies the effectiveness of our proposed pilot update mechanism. Moreover, \baby achieves very competitive results on image classification as well, as described in Section~\ref{appendix:cvres}.

\section{Analysis}


\begin{figure*}
\centering
\begin{minipage}[t]{.305\textwidth}
  \centering
  \includegraphics[width=0.9\linewidth]{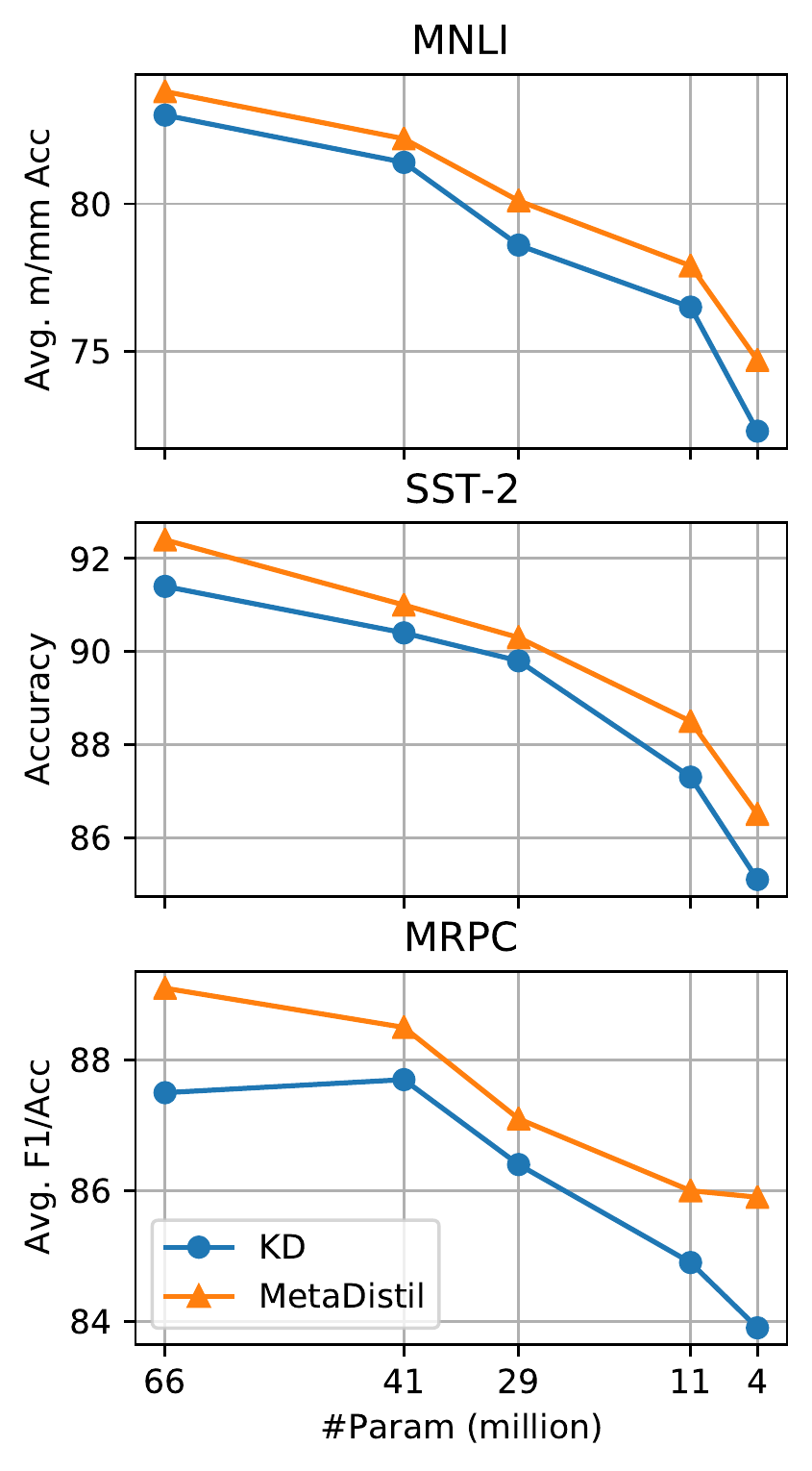}
  \captionof{figure}{Results with different student architectures.}
  \label{fig:stu}
\end{minipage}%
\hfill
\begin{minipage}[t]{.316\textwidth}
  \centering
  \includegraphics[width=0.9\linewidth]{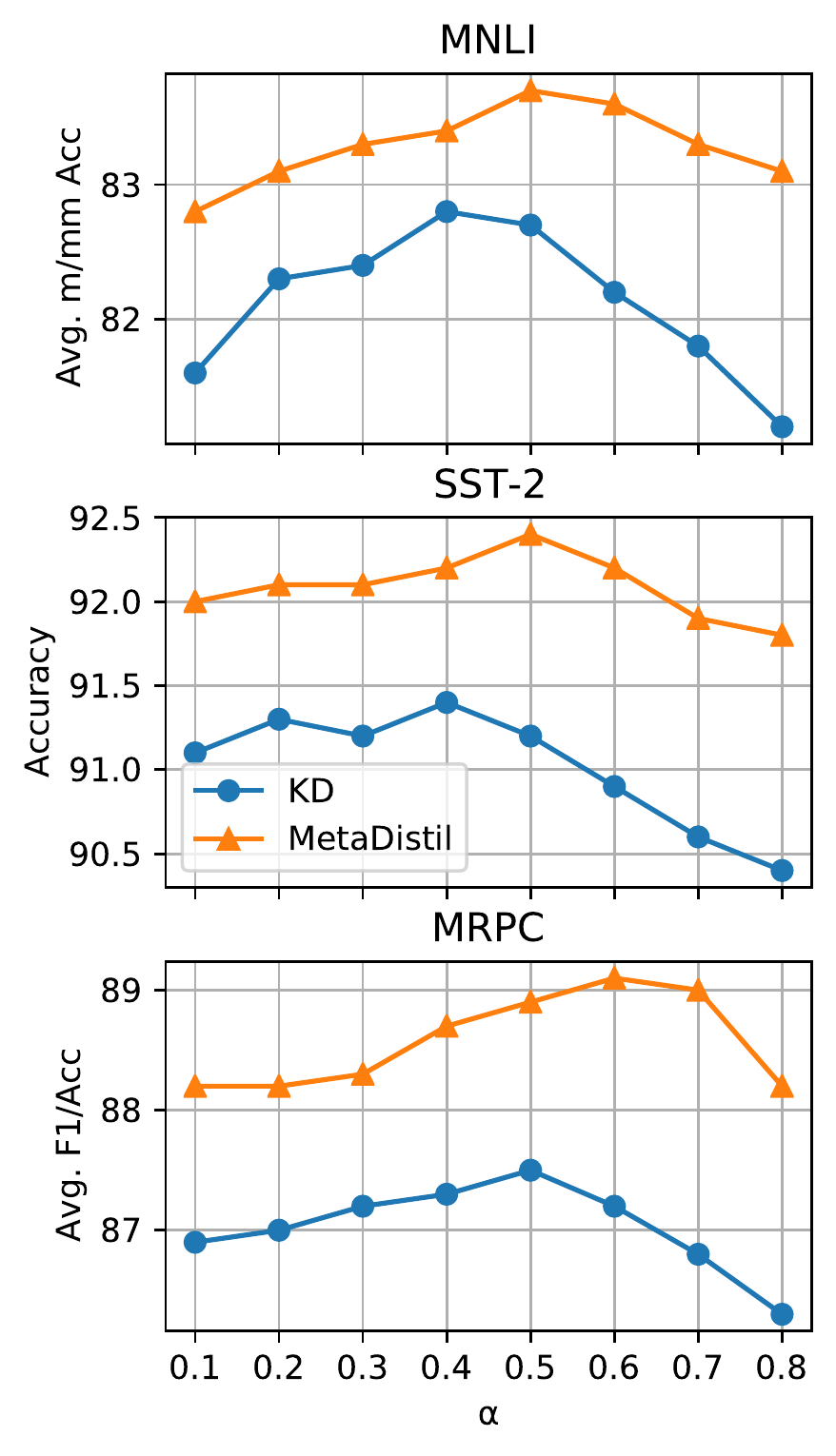}
  \captionof{figure}{Results with different loss weight $\alpha$.}
  \label{fig:weight}
\end{minipage}
\hfill
\begin{minipage}[t]{.316\textwidth}
  \centering
  \includegraphics[width=0.9\linewidth]{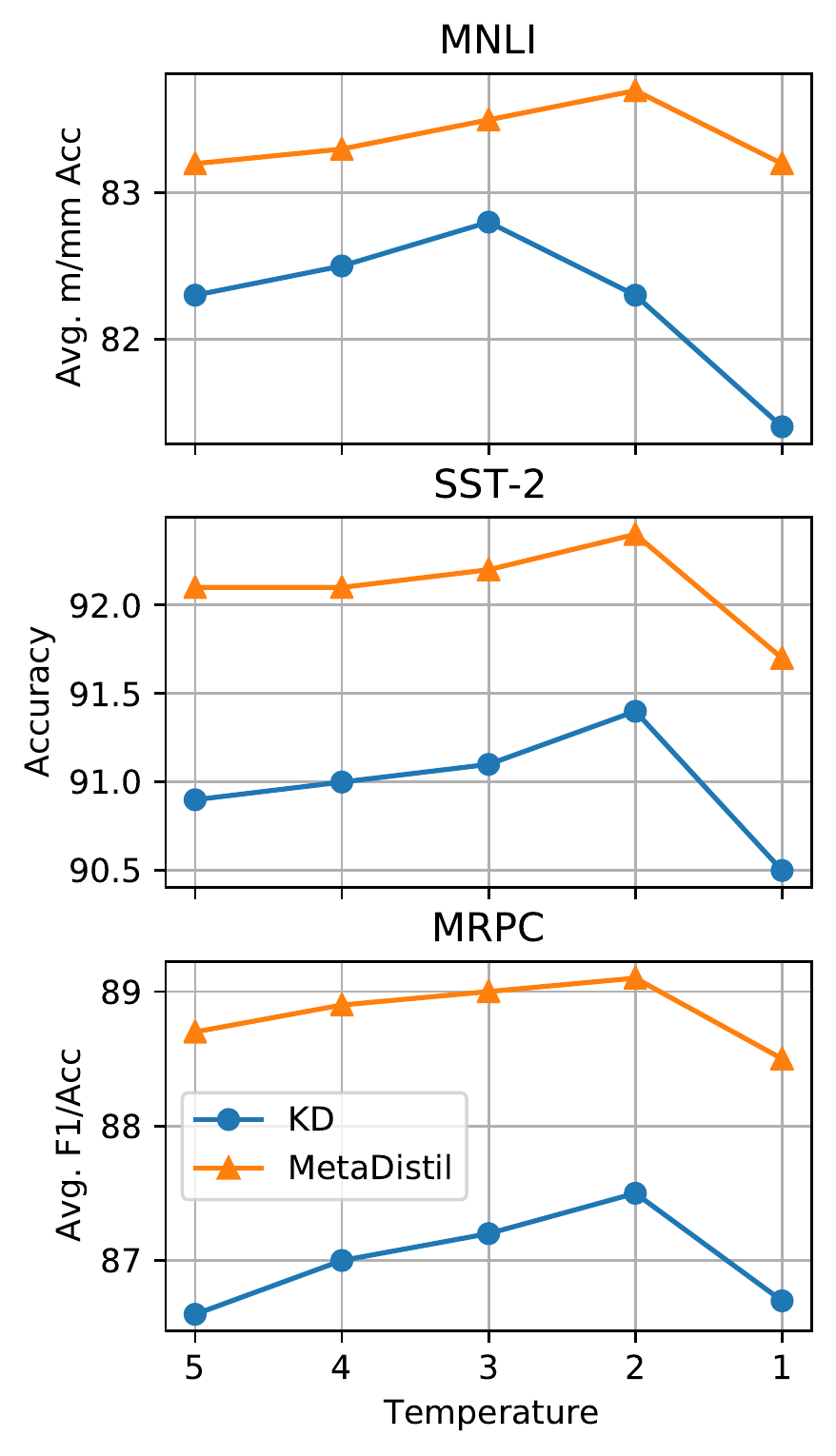}
  \captionof{figure}{Results with different temperature.}
  \label{fig:temp}
\end{minipage}
\end{figure*}

\subsection{Why Does \baby Work?}
%
%
%
%
%
%
%
%
%


We investigate the effect of meta-update for each iteration. We inspect (1) the validation loss of $S'$ after the teaching experiment and that of $S$ after the real distillation update, and (2) the KD loss, which describes the discrepancy between student and teacher, before and after the teacher update.

We find that for 87\% of updates, the student model's validation loss after real update (Line 7 in Algorithm~\ref{alg:maml}) is smaller than that after the teaching experiment (Line 4 in Algorithm~\ref{alg:maml}), which would be the update to the student $S$ in the variant without pilot update. This confirms the effectiveness of the pilot update mechanism on better matching the student and teacher model. 

Moreover, we find that
in 91\% of the first half of the updates, the teacher becomes more similar (in terms of logits distributions) to the student after the meta-update, which indicates that the teacher is learning to adapt to a low-performance student (like an elementary school teacher). However, in the second half of \baby, this percentage drops to 63\%. We suspect this is because in the later training stages, the teacher needs to actively evolve itself beyond the student to guide the student towards further improvement (like a university professor).

Finally, we try to apply a meta-learned teacher to a conventional static distillation and also to an unfamiliar student. We describe the results in details in Section~\ref{appendix:stct}.

\subsection{Hyper-parameter Sensitivity}
A motivation of \baby is to enable the teacher to dynamically adjust its knowledge transfer in an optimal way. Similar to Adam~\citep{adam} vs.~SGD~\citep{sgd1,sgd2} for optimization, with the ability of dynamic adjusting, it is natural to expect \baby to be more insensitive and robust to changes of the settings. Here, we evaluate the performance of \baby with students of various capability, and a wide variety of hyperparameters, including \textit{loss weight} and \textit{temperature}.

\paratitle{Student Capability} To investigate the performance of \baby under different student capacity, we experiment to distill BERT-Base into BERT-6L, Medium, Small, Mini and Tiny~\citep{turc2019well} with conventional KD and \baby. We plot the performance with the student's parameter number in Figure~\ref{fig:stu}. 
Additionally, we show results for different compression ratio in Appendix~\ref{sec:ratio}.

%
%
%
%
%

%

\paratitle{Loss Weight}
In KD, tuning the loss weight is 
non-trivial 
and often requires hyperparameter search. To test the robustness of \baby under different loss weights, we run experiments with different $\alpha$ (Equation~\ref{eq:loss}). As shown in Figure~\ref{fig:weight}, 
\baby consistently outperforms conventional KD and is less sensitive to different $\alpha$.

\paratitle{Temperature} Temperature is a re-scaling trick introduced in \citet{kd}. We try different temperatures and illustrate the performance of KD and \baby in Figure~\ref{fig:temp}. \baby shows better performance and robustness compared to KD.

\subsection{Limitation}
Like all meta learning algorithms, \baby inevitably requires two rounds of updates involving both first and second order derivatives. Thus, \baby requires additional computational time and memory than a normal KD method, which can be a limitation of our method. We compare the computational overheads of \baby with other methods in Table~\ref{tab:overhead}. Although our approach takes more time to achieve its own peak performance, it can match up the performance of PKD~\citep{sun2019patient} with a similar time cost. The memory use of our method is higher than PKD and ProKT~\citep{shi2021learning}. However, this one-off investment can lead to a better student model for inference, thus can be worthy.

\begin{table}[t]
\begin{center}
\resizebox{1\linewidth}{!}{
\begin{tabular}{llll}
\toprule
  \textbf{Method} & PKD~\citey{sun2019patient} & ProKT~\citey{shi2021learning} & MetaDistil \textit{(ours)}\\
\midrule
Training Time (Best) & 13 min. & 25 min. & 31 min. \\
Training Time (Match) & 13 min. & 18 min. & 16 min. \\
Memory Cost & 4.2 GB & 6.8 GB & 11.4 GB \\
\midrule
Best Acc/F1 & 89.4/84.7 & 90.7/86.3 & 91.1/86.8 \\
\bottomrule
\end{tabular}
}
\end{center}
\caption{Comparison of training time and memory cost of MetaDistil with the baselines. ``Training Time (Best)'' denotes the training time for each method to achieve its own best performance on the development set. ``Training Time (Match)'' denotes the training time for each method to match the best performance of PKD on the development set. The batch size is 4.
All experiments are conducted on a single Nvidia V100 GPU.}
\label{tab:overhead}
\end{table}

\section{Discussion}
In this paper, we present \baby, a knowledge distillation algorithm powered by meta learning that explicitly optimizes the teacher network to better transfer its knowledge to the student network. The extensive experiments verify the effectiveness and robustness of \baby.

\section*{Ethical Consideration}

\baby focuses on improving the performance of knowledge distillation and does not introduce extra ethical concerns compared to vanilla KD methods. Nevertheless, we would like to point out that as suggested by~\citet{DBLP:journals/corr/abs-2010-03058}, model compression may lead to biases. However, this is not an outstanding problem of our method but a common risk in model compression, which needs to be addressed in the future.

\section*{Acknowledgments}
We would like to thank the anonymous reviewers and the area chair for their insightful comments. This project is partly supported by NSF Award \#1750063.

\bibliography{iclr2022_conference}
\bibliographystyle{iclr2022_conference}

\appendix
\section{\baby for Image Classification}
\label{appendix:cv}
In addition to BERT compression, we also provide results on image classification. Also, we conduct experiments of static teaching and cross teaching, to further verify the effectiveness of \baby of adapting to different students.

\subsection{Experimental Settings}
For CV, following the settings in \citet{tian2019crd}, we experiment with the image classification task on CIFAR-100~\citep{krizhevsky2009learning} with student-teacher combinations of different capacity and architectures, including ResNet~\citep{he2016deep} and VGG~\citep{vgg}. Additionally, we run a distillation experiment between different architectures (a ResNet teacher to a VGG student). We report the top-1 test accuracy of the compressed student networks. 
We inherit all hyperparameters from~\citet{tian2019crd} except for the teacher learning rate, which is grid searched from \{1e-4, 2e-4, 3e-4\}. We randomly split the original training set to a new training set and the quiz set by $9:1$. We use the KL loss in \citet{hinton2015distilling} as the distillation objective.
We compare our results with a state-of-the-art distillation method, CRD~\citep{tian2019crd} and other commonly used knowledge distillation methods~\citep{kd,fitnet,AT,SP,peng2019correlation,vid,RKD,pkt,ab,ft} including ProKT~\citep{shi2021learning} which has a trainable teacher.

\subsection{Image Recognition Results}
\label{appendix:cvres}
We show the experimental results of \baby distilling ResNet~\citep{he2016deep} and VGG~\citep{vgg} with five different teacher-student pairs. %
\baby achieves comparable performance to CRD~\citep{tian2019crd}, the current state-of-the-art distillation method on image classification while outperforming all other baselines with complex features and loss functions. Notably, CRD introduces additional negative sampling and contrastive training while our method achieves comparable performance without using these tricks. Additionally, we observe a substantial performance drop without pilot update, again verifying the importance of this mechanism.

\begin{table}[t]
  \centering
  \resizebox{1\linewidth}{!}{
  \begin{tabular}{lccccc}
    \toprule
    \textbf{Teacher} & ResNet-56 & ResNet-110 & ResNet-110 & VGG-13 & ResNet-50$^*$  \\
    \textbf{Student} & ResNet-20 & ResNet-20 & ResNet-32 & VGG-8 & VGG-8 \\
    \midrule
    Teacher & 72.34 & 74.31 & 74.31 & 74.64 & 79.34 \\
    Student & 69.06 & 69.06 & 71.14 & 70.36 & 70.36\\
    \midrule
    KD~\citey{kd} & 70.66 & 70.67 & 73.08 & 72.98 & 73.81 \\
    FitNet~\citey{fitnet} & 69.21 & 68.99 & 71.06 & 71.02 & 70.69 \\
    AT~\citey{AT} & 70.55 & 70.22 & 72.31 & 71.43 & 71.84 \\
    SP~\citey{SP} & 69.67 & 70.04 & 72.69 & 72.68 & 73.34 \\
    CC~\citey{peng2019correlation} & 69.63 & 69.48 & 71.48 & 70.71 & 70.25 \\
    VID~\citey{vid} & 70.38 & 70.16 & 72.61 & 71.23 & 70.30 \\
    RKD~\citey{RKD} & 69.61 & 69.25 & 71.82 & 71.48 & 71.50 \\
    PKT~\citey{pkt} & 70.34 & 70.25 & 72.61 & 72.88 & 73.01 \\
    AB~\citey{ab} & 69.47 & 69.53 & 70.98 & 70.94 & 70.65 \\
    FT~\citey{ft} & 69.84 & 70.22 & 72.37 & 70.58 & 70.29 \\
    ProKT~\citey{shi2021learning} & 70.98 & 70.74 & 72.95 & 73.03 & 73.90 \\
    CRD~\citey{tian2019crd} & \underline{71.16} & \textbf{71.46} & \textbf{73.48} & \textbf{73.94} & \underline{74.30} \\
    \baby & \bf 71.25 & \underline{71.40} & \underline{73.35} & \underline{73.65} & \bf 74.42 \\
    \ w/o pilot update  & 71.02 & 70.96 & 73.31 & 73.48 & 74.05 \\
    \bottomrule
  \end{tabular}
  }
  \caption{Experimental results
  on the test set of CIFAR-100. The best and second best results are marked with \textbf{boldface} and \underline{underline}, respectively. All baseline results except ProKT are reported in \citet{tian2019crd}. $^*$ResNet for ImageNet. Other ResNets are ResNet for CIFAR~\citep{he2016deep}.}
  \label{tab:cifar100}
\end{table}

\begin{table}[t]
    \centering
    \resizebox{1\linewidth}{!}{
    \begin{tabular}{llc}
            \toprule
            \bf Teacher & \bf Student & \bf Acc@1 \\
            \midrule
            \multirow{2}{*}{KD (ResNet-110)} & ResNet-32 (static) & 73.08 \\
            & ResNet-20 (static) & 70.67 \\
            \midrule
            \multirow{2}{*}{\baby} & ResNet-32 (dynamic) & 73.35 \\
            \multirow{2}{*}{(ResNet-110$\rightarrow$ResNet-32)}  & ResNet-32 (static) & 73.16 \\
            & ResNet-20 (static, cross) & 70.82 \\
            \midrule
            \multirow{2}{*}{\baby} & ResNet-20 (dynamic) & 71.40 \\
            \multirow{2}{*}{(ResNet-110$\rightarrow$ResNet-20)} & ResNet-20 (static) & 70.94 \\
            & ResNet-32 (static, cross) & 72.89 \\
            \bottomrule
        \end{tabular}}
    \caption{Experimental results of static teaching and cross teaching.\label{tab:static}}
\end{table}
\subsection{Static Teaching and Cross Teaching}
\label{appendix:stct}

\begin{table*}[tbh]
\begin{center}
\resizebox{1\linewidth}{!}{
\begin{tabular}{l|cc|cccccccc}
\toprule
\multirow{2}{*}{\textbf{Method}} & \multirow{2}{*}{\textbf{\#Param.}} & \multirow{2}{*}{\textbf{Speed-up}}  & \textbf{CoLA} & \textbf{MNLI} & \textbf{MRPC} & \textbf{QNLI} & \textbf{QQP} & \textbf{RTE} & \textbf{SST-2} & \textbf{STS-B} \\
& & & (8.5K) & (393K) & (3.7K) & (105K) & (364K) & (2.5K) & (67K) & (5.7K) \\
\midrule
BERT-Base (teacher)~\citep{devlin2018bert} & 110M & 1.00$\times$ & 58.9 & 84.6/84.9 & 91.6/87.6 & 91.2 & 88.5/91.4 & 71.4 & 93.0 & 90.2/89.8 \\
\midrule
BERT$_{4}$-KD$^\dagger$~\citep{kd} & 55M & 2.90$\times$ & 32.5 & 80.5/80.9 & 87.2/83.1 & 87.5 & 86.6/90.4 & 65.2 & 90.2 & 84.5/84.2  \\
BERT$_{4}$-PKD$^\dagger$~\citep{sun2019patient} & 55M & 2.90$\times$ & 34.2 & 80.9/81.3 & 87.0/82.9 & 87.7 & 86.8/90.5 & 66.1 & 90.5 & 84.3/84.0 \\
BERT$_{4}$-ProKT$^\dagger$~\citep{shi2021learning} & 55M & 2.90$\times$ & 36.6 & 81.4/81.9 & 87.6/83.5 & 88.0 & 87.1/90.5 & 66.8 & 90.7 & 85.2/85.1 \\
\baby$_{4}$ \textit{(ours)} & 55M & 2.90$\times$ & \bf 40.3 & \bf 82.4/82.7 & \bf 88.4/84.2 & \bf 88.6 & \bf 87.8/90.8 & \bf 67.8 & \bf 91.8 & \bf 86.3/86.0 \\
\bottomrule
\end{tabular}
}
\end{center}
\caption{Experimental results
on the development set of GLUE in the setting of distilling BERT-base in to BERT$_{4}$. $^\dagger$Results reported by us. All results reported by us are average performance of 3 runs with different random seeds.}
\end{table*}

\begin{table*}[tbh]
\begin{center}
\resizebox{1\linewidth}{!}{
\begin{tabular}{l|cc|cccccccc}
\toprule
\multirow{2}{*}{\textbf{Method}} & \multirow{2}{*}{\textbf{\#Param.}} & \multirow{2}{*}{\textbf{Speed-up}}  & \textbf{CoLA} & \textbf{MNLI} & \textbf{MRPC} & \textbf{QNLI} & \textbf{QQP} & \textbf{RTE} & \textbf{SST-2} & \textbf{STS-B} \\
& & & (8.5K) & (393K) & (3.7K) & (105K) & (364K) & (2.5K) & (67K) & (5.7K) \\
\midrule
BERT-Large (teacher)~\citep{devlin2018bert} & 345M & 1.00$\times$ &  71.5 & 86.5/86.7 & 92.5/88.7 & 92.5 & 89.6/91.8 & 73.4 & 94.5 & 91.2/90.6 \\
\midrule
BERT$_{6}$-KD$^\dagger$~\citep{kd} & 66M & 3.88$\times$ & 58.8 & 82.8/83.0 & 89.6/85.0 & 89.5 & 87.5/91.0 & 68.0 & 91.1 & 88.5/88.4 \\
BERT$_{6}$-PKD$^\dagger$~\citep{sun2019patient} & 66M & 3.88$\times$ & 59.2 & 82.9/83.1 & 89.9/85.4 & 89.8 & 87.9/91.1 & 67.9 & 91.5 & 88.2/88.0 \\
BERT$_{6}$-ProKT$^\dagger$~\citep{shi2021learning} & 66M & 3.88$\times$ & 59.8 & 83.2/83.4 & 91.0/86.5 & 90.0 & 88.2/91.0 & 68.8 & 91.6 & 88.7/88.5 \\
\baby$_{6}$ \textit{(ours)} & 66M & 3.88$\times$ & \bf 63.5 & \bf 83.9/84.3 & \bf 91.5/87.3 & \bf 90.8 & \bf 88.7/91.3 & \bf 70.8 & \bf 92.9 & \bf 89.6/89.4 \\
\bottomrule
\end{tabular}
}
\end{center}
\caption{Experimental results
on the development set of GLUE in the setting of distilling BERT-large in to BERT$_{6}$. $^\dagger$Results reported by us. All results reported by us are average performance of 3 runs with different random seeds.}
\end{table*}

In
\baby, the student is trained in a dynamic manner.
To investigate the effect of such a dynamic distillation process, we attempt to use the teacher at the end of \baby training to perform a static conventional KD, to verify the effectiveness of our dynamic distillation strategy. As shown in Table~\ref{tab:static}, on both experiments, 
dynamic \baby outperforms 
conventional KD and 
static distillation with the teacher at the end of 
\baby training. 

As mentioned in Section~\ref{sec:method}, a meta teacher is optimized to transfer its knowledge to a specific student network. To justify this motivation, we conduct experiments using a teacher optimized for the ResNet-32 student to statically distill to the ResNet-20 student, and also in reverse. As shown in Table~\ref{tab:static}, the cross-taught students underperform the static students taught by their own teachers by 0.27 and 0.12 for ResNet-32 and ResNet-20, respectively. This confirms our motivation that the meta teacher in \baby can adjust itself according to its student.

\section{Results of Different Compression Ratios}
\label{sec:ratio}

In this section, we present additional experimental results in settings with different compression ratios to further demonstrate the effectiveness of MetaDistil on bridging the gap between the student and teacher capacity. Specifically, we conduct experiments in the following two settings: (1) distilling BERT-base into a 4-layer BERT (110M$\rightarrow$52M) and (2) distilling BERT-large into a 6-layer BERT (345M$\rightarrow$66M). The results are shown in Table 4 and Table 5, respectively. We can see that MetaDistil consistently outperforms PKD and ProKT in both settings. This confirms the effectiveness of MetaDistil and also show its ability to adapt the teacher model to the student model, since the gap between teacher and student is even larger in these settings. 

\section{Distillation Dynamics}

 \begin{figure}
 \centering
     \includegraphics[width=0.75\linewidth]{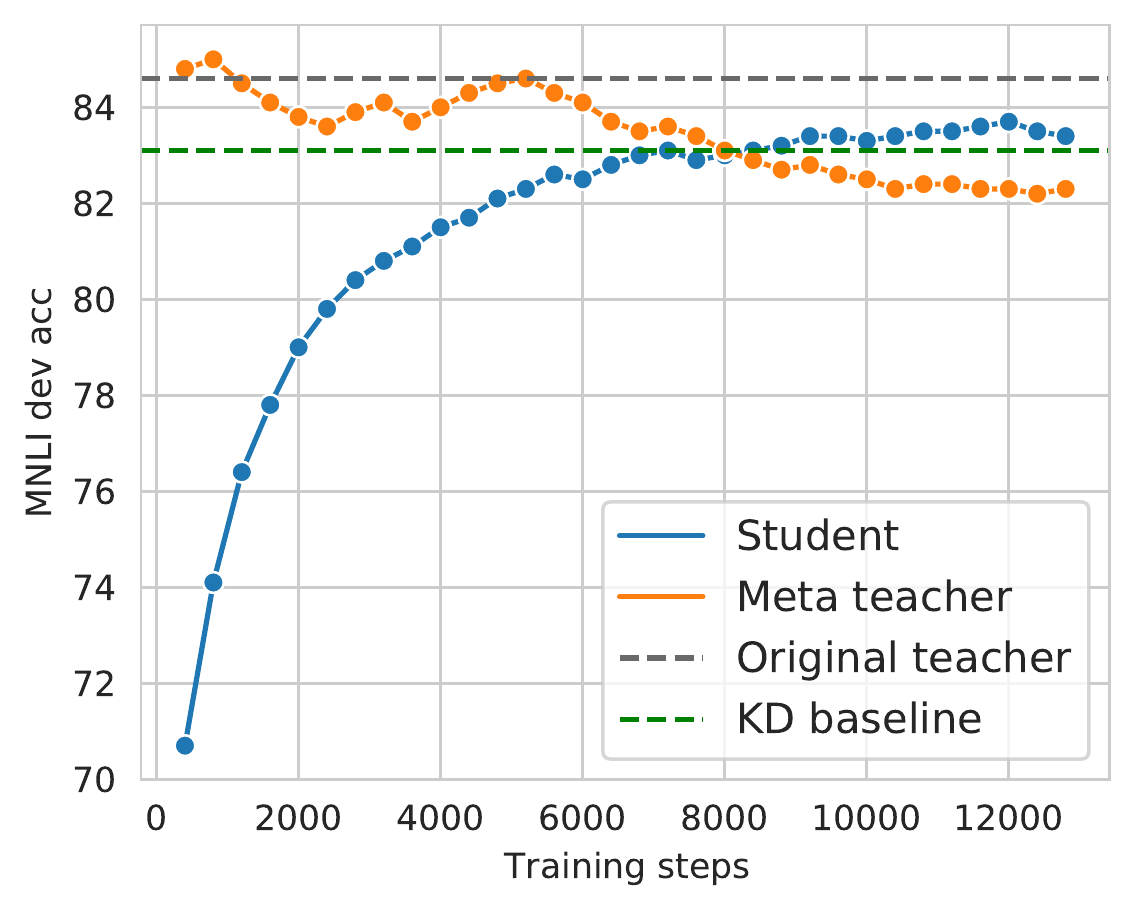}
         \captionof{figure}{Learning dynamics of the student and teacher in \baby on the development set of MNLI. \label{fig:dynamics}}
 \end{figure}

We also investigate why \baby works by conducting experiments on the development sets of MNLI, SST, and MRPC, which are important tasks in GLUE that have a large, medium, and small training set, respectively.

We illustrate the validation accuracy curves of the meta teacher and student models with training steps in Figure \ref{fig:dynamics}, and compare them to the student performance in conventional KD. We can see that the meta teacher maintains high accuracy in the first 5,000 steps and then begins to slowly degrade. Starting from step 8,000, the teacher model underperforms the student while the student's accuracy keeps increasing. This verifies our assumption that a model with the best accuracy is not necessarily the optimal teacher. Also, \baby is not naively optimizing the teacher's accuracy but its ``teaching skills.'' This phenomenon suggests that beyond high accuracy, there could be more important properties of a good teacher that warrant further investigation.

\section{Improvement Analysis}

While \baby achieves improved student accuracy on the GLUE benchmark, it is still not very clear where the performance improvement comes from. There are two possibilities: (1) the student better mimics the teacher, and (2) the changes of teacher helps student perform better on hard examples that would be incorrectly classified by the student with vanilla KD. We conduct a series of analysis on the MRPC dataset.

For the first assumption, we compute the prediction loyalty~\citep{DBLP:conf/emnlp/XuZG0MW21} of the student model distilled with PKD and \baby, respectively. For \baby, we measure the loyalty with respect to both the original teacher and the final teacher. We find that there is no significant difference between between PKD and \baby. This suggests that the improvement does not come from student better mimicking the teacher.

For the second assumption, we first identify the examples in the quiz set for which our model gives correct predictions while the student distilled by PKD makes a wrong prediction. We then compute the loss (cross entropy) of the original teacher and the teacher updated by \baby. We find the loss is substantially reduced by \baby. In contrast, the overall loss of teacher on the development set does not decrease. This suggests that \baby can help the teacher concentrate on hard examples that the student struggles in the quiz set and learn to perform better on these examples, thus facilitate student learning.

\end{document}